# Policy Iteration for Relational MDPs

**Chenggang Wang** and **Roni Khardon**
Department of Computer Science
Tufts University
Medford, MA 02155, USA
{cwan|roni}@cs.tufts.edu

## Abstract

Relational Markov Decision Processes are a useful abstraction for complex reinforcement learning problems and stochastic planning problems. Recent work developed representation schemes and algorithms for planning in such problems using the value iteration algorithm. However, exact versions of more complex algorithms, including policy iteration, have not been developed or analyzed. The paper investigates this potential and makes several contributions. First we observe two anomalies for relational representations showing that the value of some policies is not well defined or cannot be calculated for restricted representation schemes used in the literature. On the other hand, we develop a variant of policy iteration that can get around these anomalies. The algorithm includes an aspect of policy improvement in the process of policy evaluation and thus differs from the original algorithm. We show that despite this difference the algorithm converges to the optimal policy.

## 1 Introduction

Relational Markov Decision Processes (RMDP) are a useful abstraction for complex reinforcement learning problems and stochastic planning problems, since one can develop abstract solutions for them that are independent of domain size or instantiation. Propositional solutions, obtained by grounding a domain instance, do not provide such abstract solutions and they get slower with increasing problem size. Recent work on RMDPs includes exact solutions with value iteration (Boutilier et al., 2001; Kersting et al., 2004; Hölldobler & Skvortsova, 2004; Wang et al., 2007), and several approximate or heuristic methods (Guestrin et al., 2003; Fern et al., 2003; Gretton & Thiebaux, 2004; Sanner & Boutilier, 2005; Sanner & Boutilier, 2006) using different representation languages and algorithms. Motivated by the success of using decision diagrams for propositionally factored MDPs, we have previously introduced First Order Decision Diagrams (FODD) and developed a value iteration algorithm working with this representation (Wang et al., 2007).

It is interesting that all exact solution methods listed above use the value iteration algorithm. Other algorithms and in particular Policy Iteration may have obvious advantages, especially in cases where the abstract value function requires an infinite number of state partitions but there is a simple optimal policy for the domain. Kersting et al. (2004) have shown that this happens even in a simple version of the blocks world problem.

In this paper we investigate the potential of developing and analyzing policy iteration for relational domains in the context of FODDs. We introduce a new algorithm, Relational Modified Policy Iteration, that uses special operations with FODDs to mimic the original modified policy iteration algorithm. In the process we point out two anomalies of policy languages in the context of policy evaluation. First, some policy languages in the literature do not have well defined value functions. Second, there is some interaction between the value and policy languages so that, when using a restricted representation scheme, the value function of some natural policies are not expressible in the language. Overcoming this difficulty, our algorithm incorporates an aspect of policy improvement into policy evaluation. We show that the algorithm converges to the optimal value function and policy.

The rest of the paper is organized as follows. The next two sections provide background on MDPs and the value iteration solution with FODDs. Sections 4 and 5 provide the algorithm and its analysis respectively. The final section concludes with open questions raised by this work.

## 2 Markov Decision Processes

We assume familiarity with standard notions of MDPs and commonly used solution methods (Puterman, 1994). In the following we introduce some of the notions and our notation.

A MDP can be characterized by a state space $S$, an action space $A$, a state transition function $Pr(s_j|s_i, a)$ denoting the probability of transition to state $s_j$ given state $s_i$ and action $a$, and an immediate reward function $r(s)$, specifying the immediate utility of being in state $s$. A solution to a MDP is an optimal policy that maximizes expected discounted total reward as defined by the Bellman equation. The value iteration algorithm (VI) uses the Bellman equation to iteratively refine an estimate of the value function:

$$V_{n+1}(s) = max_{a \in A}[r(s) + \gamma \sum_{s' \in S} Pr(s'|s, a)V_n(s')] \quad (1)$$

where $V_n(s)$ represents our current estimate of the value function and $V_{n+1}(s)$ is the next estimate. The set of actions $\pi_{n+1}(s) = argmax_{a \in A}[r(s) + \gamma \sum_{s' \in S} Pr(s'|s, a)V_n(s')]$ are called the greedy policy with respect to the value function $V_n$. Since we can calculate $V_{n+1}$ and $\pi_{n+1}$ at the same time, we introduce the notation $(V_{n+1}, \pi_{n+1}) = \text{greedy}(V_n)$.

Similarly, for any value function $V$ and policy $\pi$ we can define

$$Q_V^\pi(s) = r(s) + \gamma \sum_{s' \in S} Pr(s'|s, \pi(s))V(s')$$

so that $Q_V^\pi$ is the value obtained by executing policy $\pi$ for one step and then receiving the terminal value $V$. When $V$ is $V_n$ and the output is taken as $V_{n+1}$ we get the successive approximation algorithm (Puterman, 1994) calculating the value function $V^\pi$ that is associated with the policy $\pi$. We later distinguish the algorithm calculating $Q$ from the actual $Q$ value in order to check whether they are the same in the relational case. We therefore denote the algorithm calculating $Q$ from $V$ and $\pi$ by regress-policy$(V, \pi)$.

Policy Iteration (PI) is an alternative algorithm that can be faster than VI. For computational efficiency, Puterman (1994) introduced Modified Policy Iteration (MPI) where the sequence $m_n$ of non-negative integers controls the number of updates in policy evaluation steps:

**Procedure 1** *Modified Policy Iteration*

1. *$n = 0$, $V_0 = R$.*
2. *Repeat*
    (a) *(Policy improvement)*
        $(w_{n+1}^0, \pi) = greedy(V_n)$.
    (b) *If $\|w_{n+1}^0 - V_n\| \leq \epsilon(1 - \gamma)/2\gamma$, return $V_n$ and $\pi$, else go to step 2c.*
    (c) *(Partial policy evaluation)*
        $k=0$.
        *while $k < m_{n+1}$ do*
        i. $w_{n+1}^{k+1} = regress\text{-}policy(w_{n+1}^k, \pi)$.
        ii. $k=k+1$.
    (d) $V_{n+1} = w_{n+1}^{m_{n+1}}$, $n = n+1$.

## 3 First Order Decision Diagrams and Value Iteration

In this section we briefly review the representation we use — first order decision diagrams. We only present the parts needed in our construction and refer the reader to (Wang et al., 2007) for further details.

A First Order Decision Diagram (FODD) is a labeled directed acyclic graph, where each non-leaf node has exactly two children. The outgoing edges are marked with values `true` and `false`. Each non-leaf node is labeled with an atom $P(t_1, \ldots, t_n)$ or an equality $t_1 = t_2$ where $t_i$ is a variable or a constant. Leaves are labeled with numerical values.

As for first order logical formulas, the semantics of FODDs is given relative to interpretations. An interpretation has a domain of elements, a mapping of constants to domain elements, and for each predicate a relation over the domain elements which specifies when the predicate is true. The semantics is defined first relative to a variable valuation $\zeta$. Given a FODD $B$ over variables $\vec{x}$ and an interpretation $I$, a valuation $\zeta$ maps each variable in $\vec{x}$ to a domain element in $I$. Once this is done, each node predicate evaluates either to `true` or `false` and we can traverse a single path to a leaf. The value of this leaf is denoted by $\text{MAP}_B(I, \zeta)$. The value of the diagram on the interpretation is defined by aggregating over all reachable leaves, and here we use maximum aggregation, that is $\text{MAP}_B(I) = max_\zeta\{\text{MAP}_B(I, \zeta)\}$. This corresponds to existential quantification in the binary case, and gives useful maximization for value functions in the general case.

Figure 1 shows a FODD with binary leaves. Left going edges represent `true` branches. To simplify diagrams in the paper we draw multiple copies of the leaves 0 and 1 but they represent the same node in the FODD. Consider evaluating the diagram in Figure 1 on the interpretation $I$ with domain $\{1, 2, 3\}$ and relations $\{p(1), q(2), h(3)\}$. The valuation $\{x/1\}$ leads to a leaf with value 0 but the valuation $\{x/2, y/3\}$ leads to a leaf with value 1 so the final result is 1.

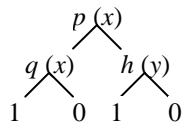

Figure 1: A simple FODD.

As in propositional algebraic decision diagrams (Bahar et al., 1993) we use a a total order over node labels, and we have a set of reduction operators to minimize representation size. Due to space constraints we omit these details that can be found in (Wang et al., 2007).

Following Boutilier et al. (2001) we specify stochastic actions as a randomized choice among deterministic alternatives. The domain dynamics of deterministic action alternatives are defined by *truth value diagrams* (TVDs). For every action schema $A(\vec{a})$ and each predicate schema $p(\vec{x})$ the TVD $T(A(\vec{a}), p(\vec{x}))$ is a FODD with $\{0, 1\}$ leaves. The TVD gives the truth value of $p(\vec{x})$ in the next state when $A(\vec{a})$ has been performed in the current state. We call $\vec{a}$ action parameters, and $\vec{x}$ predicate parameters. No other variables are allowed in the TVD. Probabilities, rewards, and value functions can be represented directly using FODDs, with the restriction that probability FODDs are either propositional or depend directly on the action parameters (Wang et al., 2007).

We use the following simple domain to illustrate the representation and later also the MPI algorithm. The domain is deterministic but it is sufficient to demonstrate the crucial representational issues. The domain includes four predicates: $p_1(x), p_2(x), q_1(x), q_2(x)$ and three deterministic actions $A_1$, $A_2$, and no-op, where $A_1(x)$ makes $p_1(x)$ true if $q_1(x)$ is true, and $A_2(x)$ makes $p_2(x)$ true if $q_2(x)$ is true. The action "no-op" does not change anything. The reward function, capturing a planning goal, awards a reward of 10 if the formula $\exists x, p_1(x) \wedge p_2(x)$ is true. We assume the discount factor is 0.9 and that there is an absorbing state $\exists x, p_1(x) \wedge p_2(x)$, i.e., no extra value will be gained once the goal is satisfied. Notice that this is indeed a very simple domain. The optimal policy needs at most two steps using $A_1$ and $A_2$ to bring about $p_1()$ and $p_2()$ (if that is possible) to reach the goal. Figure 2(a) gives the reward function for this domain and TVDs are given in Figure 2(b)(c). All the TVDs omitted in the figure are trivial in the sense that the predicate is not affected by the action.

The general first order value iteration algorithm (Boutilier et al., 2001) works as follows: given the reward function $R$ and the action model as input, we set $V_0 = R, n = 0$ and repeat the procedure *Rel-greedy* until termination:

**Procedure 2** *Rel-greedy*
 1. For each action type $A(\vec{x})$, compute:

$$Q_{V_n}^{A(\vec{x})} = R \oplus [\gamma \otimes \oplus_j (prob(A_j(\vec{x})) \otimes Regr(V_n, A_j(\vec{x})))] \quad (2)$$

 2. $Q_{V_n}^A = obj\text{-}max(Q_{V_n}^{A(\vec{x})})$.
 3. $V_{n+1} = \max_A Q_{V_n}^A$.

In the first step we calculate the $Q$-function of a stochastic action $A(\vec{x})$ parameterized with free variables $\vec{x}$. We can regress $V_n$ through a deterministic action choice $A_j(\vec{x})$ by *block replacement*, i.e., replacing each node in $V_n$ with the corresponding TVD, with outgoing edges connected to the 0, 1 leaves of the TVD. Figure 2(d) illustrates regression by *block replacement*. As discussed in (Wang et al., 2007) block replacement may not be efficient since it will necessitate reordering nodes and reductions. An alternative bottom up method *block combination* can calculate the same diagram more efficiently. We omit these details due to space constraints, but they do not affect the results in this paper. To handle the absorbing state when we have a goal based domain and have only one non-zero leaf in $R$ (as in our example), we can replace step 1 with $Q_{V_n}^{A(\vec{x})} = max(R, \gamma \otimes \oplus_j (prob(A_j(\vec{x})) \otimes Regr(V_n, A_j(\vec{x}))))$. Note that, due to discounting, the max value is always $\leq R$. If $R$ is satisfied in a state we do not care about the action (max would be $R$) and if $R$ is 0 in a state we get the value of the discounted future reward. Figure 2(e) and (f) show parameterized $Q$-functions $Q_{V_0}^{A_1(x^*)}$ and $Q_{V_0}^{A_2(x^*)}$ as calculated in the first step. The $Q$-function for *no-op* is the same as $V_0$ since this action causes no change.

In the second step we maximize over the action parameters of each $Q$-function to get the maximum value that can be achieved by using an instance of the action. With FODDs we get maximization over action parameters for free. We simply rename the action parameters using new variables names so that maximum aggregation semantics picks the correct instantiation. The third step maximizes over the $Q$-functions. All the operations among diagrams (e.g. adding) are done using a dynamic programming algorithm as in the propositional case. We have shown (Wang et al., 2007) that this algorithm correctly calculates regression and hence VI for relational domains.

## 4 Policy Iteration with FODDs

There is more than one way to represent policies with FODDs. Here we focus on one particular choice where each leaf is annotated both with a value and with a parameterized action. Note that the values are necessary; a policy with only actions at leaves is not well defined because our semantics does not support state partitions explicitly. Given a state, multiple paths may be traversed, but we only choose an action associated with the maximum value. Figure 2(g) gives an example of such a policy.

Notice that we can extend Procedure 2 to calculate the greedy policy at the same time it calculates the updated value function. To perform this we simply annotate leaves of the original $Q$-functions with the corresponding action schemas. We then perform the rest of the procedure while maintaining the action annotation. We use the notation $(V_{n+1}, \pi) = $ Rel-greedy$(V_n)$ to capture this extension where $V_{n+1}$ and $\pi$ refer to the same FODD but $V_{n+1}$ ignores the action information. The example in Figure 2(g) shows the policy that is greedy with respect to the reward function $R$. We have the following lemma:

**Lemma 1 (Wang et al., 2007)**
Let $(V_{n+1}, \pi) = $ Rel-greedy$(V_n)$, then $V_{n+1} = Q^\pi_{V_n}$.

### 4.1 The Value Associated with a Policy

In any particular state we evaluate the policy FODD and choose a binding that leads to a maximizing leaf to pick the action. However, note that if multiple bindings reach the same leaf then the choice among them is not specified. Thus our policies are not fully specified. The same is true for most of the work in this area where action choice in policies is based on existential conditions although in practical implementations one typically chooses randomly among the ground actions. Some under-specified policies are problematic since it is not clear how to define their value — the value depends on the unspecified portion, i.e., the choice of action among bindings. For example, in the blocks world where the goal is $Clear(a)$ and the policy is $Clear(x) \rightarrow MoveToTable(x)$. If $x$ is substituted with a block above $a$, then it is good; otherwise it will not help reach the goal. We therefore have:

**Observation 1** *Existential relational policies are not fully specified. As a result, their value functions may not be well defined.*

Our algorithm below does not resolve this issue. However, we show that the policies we produce have well defined values.

### 4.2 Relational Modified Policy Iteration

Relational Modified Policy Iteration (RMPI) uses the MPI procedure from Section 2 with the following three changes. 1) We replace Step 2a with $(w^0_{n+1}, \pi^0_{n+1}) = $ Rel-greedy$(V_n)$. 2) We replace the procedure regress-policy with a relational counterpart which is defined next. 3) We replace Step 2(c)i with $(w^{k+1}_{n+1}, \pi^{k+1}_{n+1}) = $ Rel-regress-policy$(w^k_{n+1}, \pi^k_{n+1})$. Notice that unlike the original MPI we change the policy which is regressed in every step.

Our relational policy regression generalizes an algorithm from (Boutilier et al., 2000) where propositional decision trees were used to solve factored MDPs.

**Procedure 3** *Rel-regress-policy*
Input: $w^i, \pi$
Output: $w^{i+1}, \hat{\pi}$
1. For each action $A()$ occurring in $\pi$, calculate the $Q$-function for the action type $A(\vec{x})$, $Q^{A(\vec{x})}_{w^i}$, using Equation (2).
2. At each leaf of $\pi$ annotated by $A(\vec{y})$, delete the leaf label, and append $Q^{A(\vec{x})}_{w^i}$ after (a) substituting the action parameters $\vec{x}$ in $Q^{A(\vec{x})}_{w^i}$ with $\vec{y}$, (b) standardizing apart the $Q$-function from the policy FODD except for the shared action arguments $\vec{y}$, and (c) annotating the new leaves with the action schema $A(\vec{y})$.
3. Return the FODD both as $w^{i+1}$ and as $\hat{\pi}$.

Figure 2(h) gives the partial result of *Rel-regress-policy* after appending $Q^{A_2(x^*)}_{w^0_1}$. Notice that we instantiate the action parameter $x^*$ with $x$. It is instructive to review the effect of reductions from (Wang et al., 2007) in this case to see that significant compaction can occur in the process. On the left side of $y = x$, we can replace each $y$ with $x$, therefore the node predicates are determined and only the leaf valued 9 remains. On the right side of $y = x$, the path leading to 10 will be dominated by the path $p_1(x), p_2(x) \rightarrow 10$ (not shown in the figure), therefore we can replace it with 0. Now we can see that the left side of $y = x$ dominates the other side. Since $y$ is a new variable it is free to take value equal to $x$. Therefore we can drop the equality and its right side. Recall that for our example domain actions are deterministic and have no "cascaded" effects so there is no use executing the same action twice. It is therefore not surprising that in this case $w^1_1 = $ Rel-regress-policy$(w^0_1, \pi^0_1)$ is the same as $w^0_1$. Figure 2(j) shows the value function and policy that are greedy with respect to $w^1_1$. These are also the optimal value function and policy.

As in the case of block replacement, a naive implementation as illustrated above may not be efficient since it necessitates node reordering and reductions. Instead we can use an idea similar to block combination to calculate the result as follows. For $i$'th leaf node $L_i$ in the policy, let $A(\vec{y})$ be the action attached. We replace $L_i$ with 1 and all the other leaf nodes with 0, and get a new FODD $F_i$. Multiply $F_i$ with $Q^{A(\vec{y})}_{w^i}$ and denote the result as $R_i$. We do this for each leaf node $L_j$ and calculate $R_j$. Finally add the results using $\oplus_j R_j$. Figure 2(i) illustrates this process.

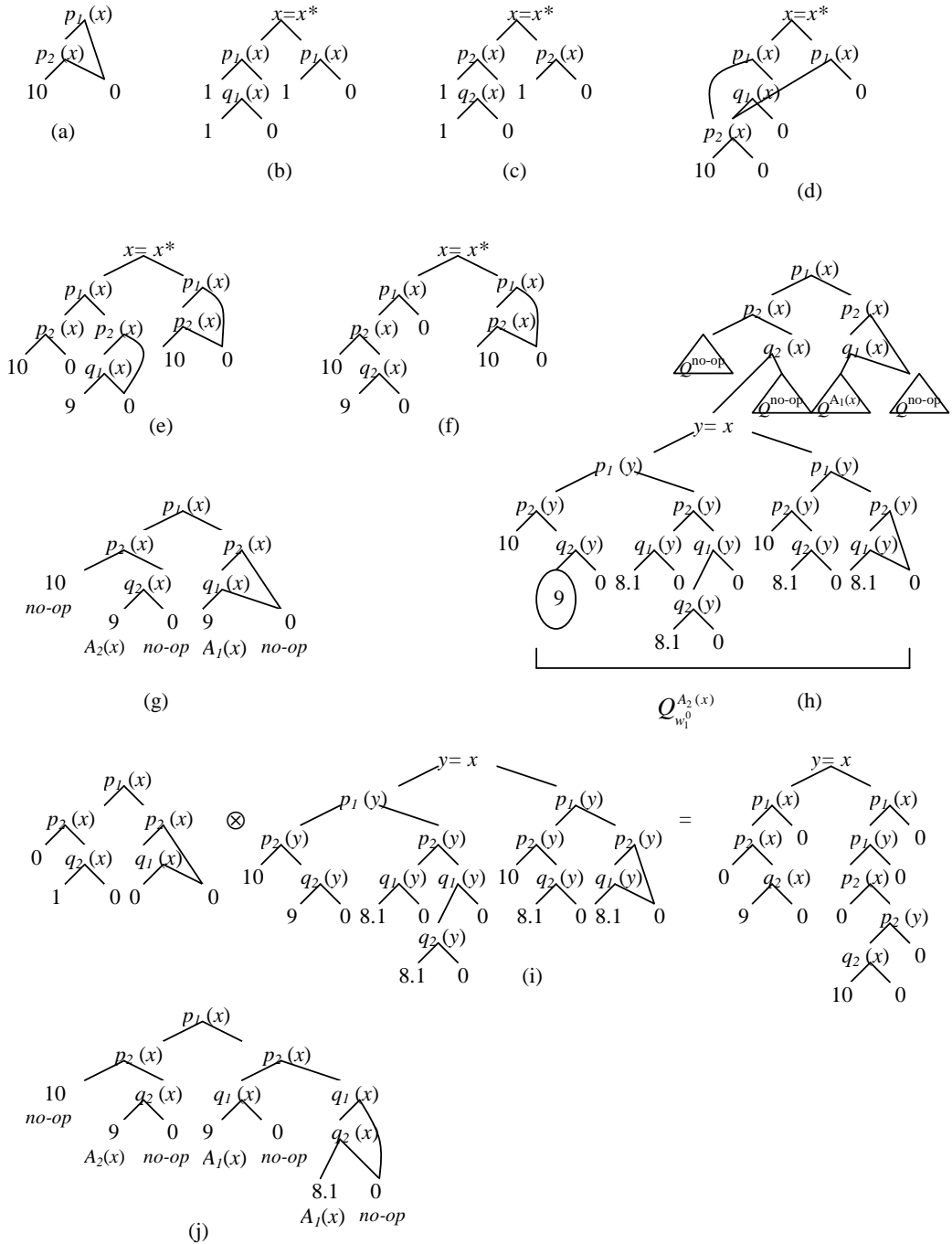

Figure 2: An example of policy iteration. (a) The reward function $R$ and the value function $V_0$. (b) The TVD for $p_1(x)$ under action $A_1(x^*)$ $T(A_1(x^*), p_1(x))$. (c) The TVD for $p_2(x)$ under action $A_2(x^*)$ $T(A_2(x^*), p_2(x))$. (d) Regression through $A_1(x^*)$ by *block replacement*. Here we replace $p_1(x)$ in $V_0$ with the TVD $T(A_1(x^*), p_1(x))$. (e) $Q_{V_0}^{A_1(x^*)}$. (f) $Q_{V_0}^{A_2(x^*)}$. (g) The value function $w_1^0$ and the policy $\pi_1^0$ such that $(w_1^0, \pi_1^0) = $ Rel-greedy($V_0$). (h) An intermediate result when performing Rel-regress-policy($w_1^0, \pi_1^0$). Note that we replace a leaf node annotated with action $A_2(x)$ with $Q_{w_1^0}^{A_2(x^*)}$ after the action parameter $x^*$ is substituted with $x$. (i) Appending $Q_{w_1^0}^{A_2(x^*)}$ through *block combination*. Reductions were used to replace the right branches of $p_1(y)$ and $p_2(y)$ with 0 in the result. (j) The value function and the policy that are greedy with respect to $w_1^1$. These are also the optimal value function and policy.

Note that all the policies produced by RMPI are calculated by Rel-greedy(V) and Rel-regress-policy($V'$, $\pi'$). Since these procedures guarantee a value achievable by leaf partitions, the maximum aggregation semantics implies that the value of our policies is well defined.

Finally, $(w^{i+1}, \hat{\pi}) = $ Rel-regress-policy($w^i, \pi$) suggests that $\hat{\pi}$ may be different from $\pi$. This is in contrast with the propositional case (Boutilier et al., 2000) where they are necessarily the same since all the leaves of a single Q-function have the same label and every interpretation follows exactly one path. In our case the policies may indeed be different and this affects the MPI algorithm. We discuss this point at length when analyzing the algorithm.

## 5　Correctness and Convergence

The procedure *Rel-regress-policy* is the key step in RMPI. The correctness of MPI relies on the correctness of the regression step. That is, the output of the regression procedure should equal $Q_V^\pi$. This is possible when one is using a representation expressive enough to define explicit state partitions that are mutually exclusive, as for example in SDP (Boutilier et al., 2001). But it may not be possible with restricted languages. In the following we show that it is not possible for implicit state partitions used in FODDs and in ReBel (Kersting et al., 2004). In particular, the regression procedure we presented above may in fact calculate an overestimate of $Q_V^\pi$. Before we expand on this observation we identify useful properties of the regression procedure.

**Lemma 2** *Let* $(w^{i+1}, \hat{\pi}) = $ *Rel-regress-policy*($w^i, \pi$), *then* $w^{i+1} = Q_{w^i}^{\hat{\pi}} \geq Q_{w^i}^\pi$.

*Proof:* $w^{i+1} = Q_{w^i}^{\hat{\pi}}$ holds by the definition of the procedure *Rel-regress-policy*.

Let $s$ be any state; we want to prove $w^{i+1}(s) \geq Q_{w^i}^\pi(s)$. We first analyze how $Q_{w^i}^\pi(s)$ can be calculated using the diagram of $w^{i+1}$. Let $A(\vec{x})$ be the maximizing action for $s$ according to the policy $\pi$, and $\zeta$ a valuation that reaches the leaf node annotated by the action $A(\vec{x})$ in $\pi$, i.e., $\text{MAP}_\pi(s, \zeta) = max_{\zeta_1}\{\text{MAP}_\pi(s, \zeta_1)\}$. Denote the part of $\zeta$ that corresponds to action parameters $\vec{x}$ as $\zeta_A$ and let $\zeta_{v_1} = \zeta \setminus \zeta_A$.

Recall that $w^{i+1}$ is obtained by replacing each leaf node of $\pi$ with the corresponding Q-function. For state $s$, $\zeta$ will reach the root node of $Q_{w^i}^{A(\vec{x})}$ in $w^{i+1}$. Note that the variables in $\pi$ and $Q_{w_i}^{A(\vec{x})}$ are all standardized apart except that they share action parameters $\vec{x}$, therefore the valuation to $\vec{x}$ in $Q_{w_i}^{A(\vec{x})}$ has been fixed by $\zeta_A$. We denote the valuation to all the other variables in $Q_{w^i}^{A(\vec{x})}$ as $\zeta_{v_2}$. The final value we get for $Q_{w^i}^\pi(s)$ is $max_{\zeta_{v_2}} Q_{w^i}^{A(\zeta_A)}$. That is, if $\eta$ is the valuation to variables in $w^{i+1}$ such that $\eta = \zeta + \zeta_{v_2}$ then we have $\text{MAP}_{w^{i+1}}(s, \eta) = Q_{w^i}^\pi(s)$. Since $w^{i+1}(s)$ is determined by the maximal value any valuation can achieve, we get $w^{i+1}(s) \geq Q_{w^i}^\pi(s)$. ∎

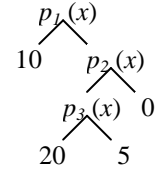

Figure 3: A reward function

We next show that the inequality in Lemma 2 may be strict, i.e., we may have $w^{i+1} = Q_{w^i}^{\hat{\pi}} > Q_{w^i}^\pi$, and this happens when $\pi \neq \hat{\pi}$. This is demonstrated by the following example. Consider a domain with reward function as shown in Figure 3, and with three actions $A_1$, $A_2$, and *no-op*. $A_1(x)$ makes $p_1(x)$ true if $q_1(x)$ is true, i.e., $q_1(x) \to p_1(x)$. $A_2(x)$ makes $p_2(x)$ true if $q_2(x)$ is true, and $p_3(x)$ true if $p_2(x)$ is true, i.e., $q_2(x) \to p_2(x)$ and $p_2(x) \to p_3(x)$. Therefore in some states it is useful to do the same action twice. Note that once $p_1(x)$ is true, there is no way to undo it. For example, if we execute action $A_1(1)$ in state $\{q_1(1), \neg p_1(1), p_2(1), \neg p_3(1)\}$, then this state will gain the value of 10 but never be able to reach a better value 20. In the following discussion we use the decision list representation where each rule corresponds to a path in the FODD and rules are ordered by value to capture maximum aggregation. We also assume $\gamma > 0.5$. The following list gives some of the partitions in the policy and the value function $(w_1^0, \pi) = $ Rel-greedy(R).

$\neg p_1(x) \land p_2(x) \land p_3(x) \to 20 + 20\gamma$(no-op).
$\neg p_1(x) \land p_2(x) \land \neg p_3(x) \to 5 + 20\gamma(A_2(x))$.
$\neg p_1(x) \land q_2(x) \land \neg p_2(x) \land p_3(x) \to 20\gamma(A_2(x))$.
$p_1(x) \to 10 + 10\gamma$(no-op).
$q_1(x) \to 10\gamma(A_1(x))$.
$\neg p_1(x) \land q_2(x) \land \neg p_2(x) \land \neg p_3(x) \to 5\gamma(A_2(x))$.

Consider next the calculation of $(w_1^1, \hat{\pi}) = $ Rel-regress-policy($w_1^0, \pi$). To illustrate the result consider the last state partition $\neg p_1(x) \land q_2(x) \land \neg p_2(x) \land \neg p_3(x)$. After executing $A_2(x)$, this state will transition to the state $\neg p_1(x) \land p_2(x) \land \neg p_3(x)$ that is associated with the value $5 + 20\gamma$. Incorporating the discount factor we see that the regressed value for $\neg p_1(x) \land q_2(x) \land \neg p_2(x) \land \neg p_3(x)$ is $5\gamma + 20\gamma^2$. The following list gives some of the partitions for $(w_1^1, \hat{\pi})$.

$\neg p_1(x) \land p_2(x) \land p_3(x) \to 20 + 20\gamma + 20\gamma^2$(no-op).
$\neg p_1(x) \land p_2(x) \land \neg p_3(x) \to 5 + 20\gamma + 20\gamma^2(A_2(x))$.
$\neg p_1(x) \land q_2(x) \land \neg p_2(x) \land p_3(x) \to 20\gamma + 20\gamma^2(A_2(x))$.
$p_1(x) \to 10 + 10\gamma + 10\gamma^2$(no-op).

$\neg p_1(x) \wedge q_2(x) \wedge \neg p_2(x) \wedge \neg p_3(x) \rightarrow 5\gamma + 20\gamma^2(A_2(x))$.
$q_1(x) \rightarrow 10\gamma + 10\gamma^2(A_1(x))$.

Note that state partitions $\neg p_1(x) \wedge q_2(x) \wedge \neg p_2(x) \wedge \neg p_3(x)$ and $q_1(x)$ have now switched places because state partitions are sorted by values in decreasing order. Therefore in this example $\pi \neq \hat{\pi}$. Now suppose we have a state $\{q_1(1), \neg p_1(1), \neg p_2(1), \neg p_3(1), q_2(1)\}$. According to $\pi$ we should choose $q_1(x) \rightarrow 10\gamma(A_1(x))$ and execute the action $A_1(1)$. But according to $\hat{\pi}$, we should follow $\neg p_1(x) \wedge q_2(x) \wedge \neg p_2(x) \wedge \neg p_3(x) \rightarrow 5\gamma + 20\gamma^2(A_2(x))$ and execute the action $A_2(1)$. This shift in policy happens because state partitions are not explicitly mutually exclusive and the choice among them is determined by their value.

Therefore, due to the use of maximum aggregation, policy evaluation with our representation incorporates an element of policy improvement. It is also important to note that this is not just a result of the procedure we use but in fact a limitation of the representation, in this case because we cannot capture universal quantification. In the example above, to represent the policy, i.e., to force executing the policy even if it does not give the maximal value for a state, we need to represent the state partition $\neg p_1(x) \wedge q_2(x) \wedge \neg p_2(x) \wedge \neg p_3(x) \wedge \neg \exists y, q_1(y) \rightarrow A_2(x)$.

The condition $\neg \exists y, q_1(y)$ must be inserted during policy evaluation to make sure that we evaluate the original policy, i.e., to execute $A_1(x)$ whenever $\exists x, q_1(x)$ is true. This involves universal quantification but our representation is not expressive enough to represent every policy. On the other hand, our representation is expressive enough to represent each iterate in value iteration as well as the optimal policy when we start with a reward function with existential quantification. This is true because each value function in the sequence includes a set of existential conditions, each showing that if the condition holds then a certain value can be achieved. From the example and discussion we have:

**Observation 2** *There exist domains and well defined existential relational policies such that (1) regress-policy cannot be expressed within an existential language, and (2) $w^{i+1} = Q_{w^i}^{\hat{\pi}} > Q_{w^i}^{\pi}$.*

The obvious question is whether our RMPI algorithm converges to the correct value and policy. If we do not change the regression policy in every step, our algorithm can be seen either as getting a wrong value in regression or as regressing for a different policy than intended ($\hat{\pi}$ instead of $\pi$) so standard convergence proofs for MPI do not apply. The standard proofs also do not apply when we do change the policy in every step. However, as we show below because $\hat{\pi}$ improves over $\pi$ w.r.t. $w_i$ the algorithm does converge. For the following analysis we have encapsulated all the FODD dependent properties in Lemma 2. The arguments hold for any representation for which Lemma 2 is true.

To facilitate the analysis we represent the sequence of value functions and policies from our algorithm as $\{(y^k, \pi^k)\}$ with $(y^0, \pi^0) = (R, \text{no-op})$. Using this notation $(y^k, \pi^k) = \text{Rel-greedy}(y^{k-1})$ when $y^k$ corresponds to $w_n^0$ for some $n$, and $(y^k, \pi^k) = \text{Rel-regress-policy}(y^{k-1}, \pi^{k-1})$ where $y^k$ corresponds to $w_n^i$ and $i > 0$.

**Lemma 3** *(a) $y^{n+1} = Q_{y^n}^{\pi^{n+1}} (n \geq 0)$.
(b) $Q_{y^n}^{\pi^{n+1}} \geq Q_{y^n}^{\pi^n}$.*

*Proof:* When $(y^{n+1}, \pi^{n+1}) = \text{Rel-greedy}(y^n)$, then by Lemma 1, $y^{n+1} = Q_{y^n}^{\pi^{n+1}}$. By the property of greedy policies, i.e. $\forall \pi', Q_{y^n}^{\pi^{n+1}} \geq Q_{y^n}^{\pi'}$ we get $Q_{y^n}^{\pi^{n+1}} \geq Q_{y^n}^{\pi^n}$. When $(y^{n+1}, \pi^{n+1}) = \text{Rel-regress-policy}(y^n, \pi^n)$, then by Lemma 2, $y^{n+1} = Q_{y^n}^{\pi^{n+1}} \geq Q_{y^n}^{\pi^n}$. ∎

Notice that our proof for Lemma 3 requires that we change the policy in every step of policy evaluation since the improvement in $Q$ value is only guaranteed relative to the policy we attempt to regress upon.

**Lemma 4** $\forall k, \exists \tilde{\pi}_k$ *such that* $V^{\tilde{\pi}_k} \geq y^k$.

*Proof:* The proof is by induction on $k$. The induction hypothesis is satisfied for k=0 because the policy that always performs *no-op* can achieve value $\geq R$. Assume it holds for $k = 1, 2, \ldots, n$, i.e., $y^k (k = 1, 2, \ldots, n)$ can be achieved by executing some policy $\tilde{\pi}_k$. By Lemma 3 $y^{n+1}$ can be achieved by acting according to $\pi^{n+1}$ in the first step, and acting according to $\tilde{\pi}_n$ in the next $n$ steps. ∎

Note that it follows from this lemma that $y^i$ is achievable by some (possibly non-stationary) policy and since stationary policies are sufficient (Puterman, 1994) we have

**Lemma 5** $y^i \leq V^*$, *where $V^*$ is the optimal value function.*

Next we prove that the value function sequence is monotonically increasing if $R \geq 0$.

**Lemma 6** $\forall i, y^{i+1} \geq y^i$.

*Proof:* We prove $y^{i+1} \geq y^i$ by induction. When $i = 0$, $(y^1, \pi^1) = \text{Rel-greedy}(y^0)$ and $y^0 = R$.

$$\begin{aligned} y^1(s) &= R(s) + \gamma \sum Pr(s'|s, \pi^1(s))y^0(s') \\ &\geq R(s) (\text{By } y^0 = R \geq 0) \\ &= y^0(s) \end{aligned}$$

When $i = 1$, $(y^2, \pi^2) = $ Rel-regress-policy$(y^1, \pi^1)$.

$$\begin{aligned} y^2(s) &= R(s) + \gamma \sum Pr(s'|s, \pi^2(s))y^1(s') \\ &\geq R(s) + \gamma \sum Pr(s'|s, \pi^1(s))y^1(s') \\ &\quad \text{(By Lemma 2)} \\ &\geq R(s) + \gamma \sum Pr(s'|s, \pi^1(s))y^0(s') \\ &\quad \text{(By the first base case)} \\ &= y^1(s) \end{aligned}$$

Assume the hypothesis holds for $i = 3, 4, \ldots, n$. Now we want to prove $y^{n+1} \geq y^n$.

$$\begin{aligned} y^{n+1}(s) &= R(s) + \gamma \sum Pr(s'|s, \pi^{n+1}(s))y^n(s') \\ &\quad \text{(By Lemma 3 (a))} \\ &\geq R(s) + \gamma \sum Pr(s'|s, \pi^n(s))y^n(s') \\ &\quad \text{(By Lemma 3 (b))} \\ &\geq R(s) + \gamma \sum Pr(s'|s, \pi^n(s))y^{n-1}(s') \\ &\quad \text{(By assumption)} \\ &= y^n(s) \end{aligned}$$

∎

The assumption $R \geq 0$ can be replaced by requiring that $w^0_{N+1} = greedy(V_N) \geq V_N$ (e.g. for $N = 0$) implying that $\forall i \geq N$, $y^{i+1} \geq y^i$. Let $V_i^{VI}$ denote each iterate in value iteration. Lemma 6 and the fact that if $v \geq u$ then $\hat{v} \geq \hat{u}$ where $\hat{v} = greedy(v)$ and $\hat{u} = greedy(u)$ (Puterman, 1994) imply:

**Lemma 7** $V_k \geq V_k^{VI}$.

Now from Lemma 5 and Lemma 7 we get:

**Theorem 1** $\{y^n\}$ *converges monotonically to the optimal value function* $V^*$.

## 6  Discussion and Future Work

The paper introduced Relational Modified Policy Iteration using FODD representations. We have observed that policy languages have an important effect on correctness and potential of policy iteration since the value of a policy may not be expressible in the language. Our algorithm overcomes this problem by including an aspect of policy improvement into policy evaluation. We have shown that the algorithm converges to the optimal value function and policy and that it dominates the iterates from value iteration.

Several interesting questions remain concerning efficiency and alternative algorithmic ideas. Some of these can become clearer through an experimental evaluation. First it would be interesting to compare the relational versions of VI and PI in terms of efficiency and convergence speed. A second issue is the stopping criterion. Pure PI stops when the policy does not change, but MPI adopts a value difference as in VI. It is possible that we can detect convergence of the policy at the structural level (ranking or aggregating actual values), and in this way get an early stopping criterion. But it is not clear how widely applicable such an approach is and whether we can guarantee correctness. Finally, RMPI uses a different policy in every step of policy evaluation. It would be interesting to analyze the algorithm if the policy is kept fixed in this process.

### Acknowledgments

We are grateful to Saket Joshi for useful discussions.

### References

Bahar, R. I., Frohm, E. A., Gaona, C. M., Hachtel, G. D., Macii, E., Pardo, A., & Somenzi, F. (1993). Algebraic decision diagrams and their applications. *Proc. of the International Conference on Computer-Aided Design*.

Boutilier, C., Dean, T., & Goldszmidt, M. (2000). Stochastic dynamic programming with factored representations. *Artificial Intelligence*, *121(1)*, 49–107.

Boutilier, C., Reiter, R., & Price, B. (2001). Symbolic dynamic programming for first-order MDPs. *Proc. of the International Joint Conference of Artificial Intelligence*.

Fern, A., Yoon, S., & Givan, R. (2003). Approximate policy iteration with a policy language bias. *International Conference on Neural Information Processing Systems*.

Gretton, C., & Thiebaux, S. (2004). Exploiting first-order regression in inductive policy selection. *Proceedings of the International Conference on Machine Learning*.

Guestrin, C., Koller, D., Gearhart, C., & Kanodia, N. (2003). Generalizing plans to new environments in relational MDPs. *Proceedings of the International Joint Conference of Artificial Intelligence*.

Hölldobler, S., & Skvortsova, O. (2004). A logic-based approach to dynamic programming. *AAAI-04 workshop on learning and planning in Markov Processes – advances and challenges*.

Kersting, K., Otterlo, M. V., & Raedt, L. D. (2004). Bellman goes relational. *Proceedings of the International Conference on Machine Learning*.

Puterman, M. L. (1994). *Markov decision processes: Discrete stochastic dynamic programming*. Wiley.

Sanner, S., & Boutilier, C. (2005). Approximate linear programming for first-order MDPs. *Proceedings of the Conference on Uncertainty in Artificial Intelligence*.

Sanner, S., & Boutilier, C. (2006). Practical linear value-approximation techniques for first-order MDPs. *Proc. of the Conference on Uncertainty in Artificial Intelligence*.

Wang, C., Joshi, S., & Khardon, R. (2007). First order decision diagrams for relational MDPs. *Proceedings of the International Joint Conference of Artificial Intelligence*.